# LatteReview☕
# A Multi-Agent Framework for Systematic Review Automation Using Large Language Models

## (A Technical Report)

**Pouria Rouzrokh[1,2,*], Bardia Khosravi[1,2], Parsa Rouzrokh[2], Moein Shariatnia[3]**

(1) Department of Radiology & Biomedical Imaging, Yale School of Medicine, Yale University, CT, USA (2) Mayo Clinic AI Laboratory, Mayo Clinic, MN, USA (3) Tehran University of Medical Sciences, Tehran, Iran

(*) Please address all correspondence to pouria.rouzrokh@yale.edu.

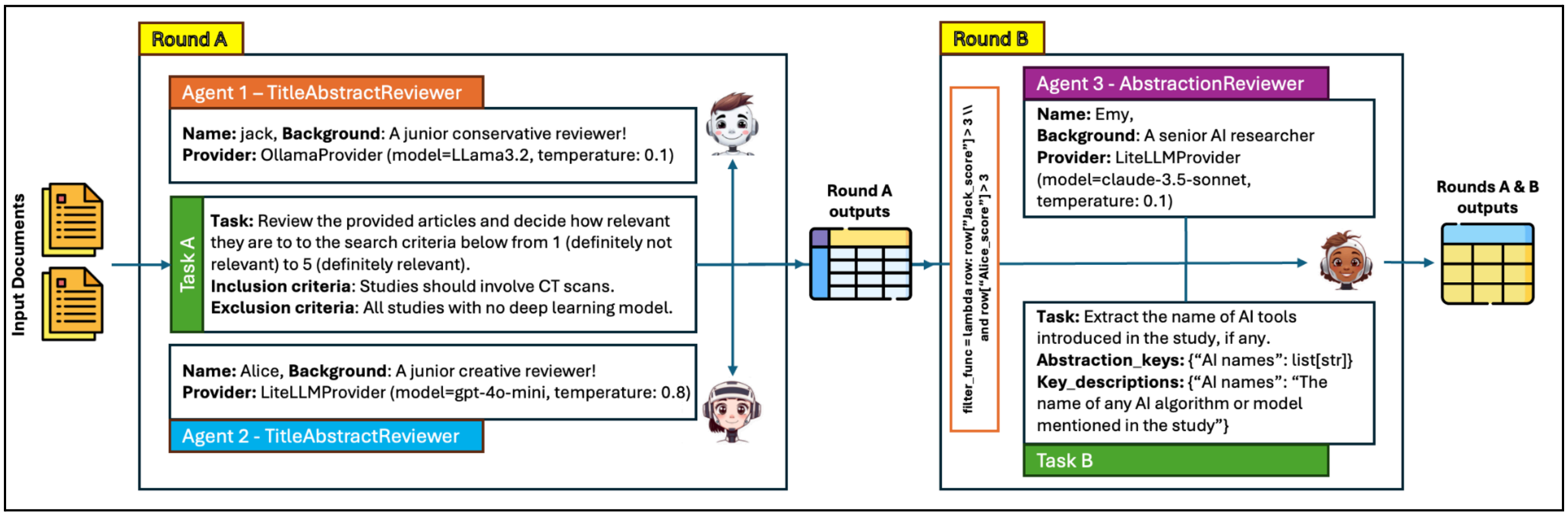

**Figure 1.** An example multiagent review workflow for title and abstract screening with two junior reviewer agents (round A) followed by concept extraction by a senior and more powerful reviewer agent (round B).

## Abstract

Systematic literature reviews and meta-analyses are essential for synthesizing research insights, but they remain time-intensive and labor-intensive due to the iterative processes of screening, evaluation, and data extraction. This paper introduces and evaluates LatteReview, a Python-based framework that leverages large language models (LLMs) and multi-agent systems to automate key elements of the systematic review process. Designed to streamline workflows while maintaining rigor, LatteReview utilizes modular agents for tasks such as title and abstract screening, relevance scoring, and structured data extraction. These agents operate within orchestrated workflows, supporting sequential and parallel review rounds, dynamic decision-making, and iterative refinement based on user feedback. LatteReview's architecture integrates LLM providers, enabling compatibility with both cloud-based and locally hosted models. The framework supports features such as Retrieval-Augmented Generation (RAG) for incorporating external context, multimodal reviews, Pydantic-based validation for structured inputs and outputs, and asynchronous programming for handling large-scale datasets. The framework is available on the GitHub repository, with detailed documentation and an installable package.

# 1. Background

Academic literature reviews are foundational to advancing knowledge in any discipline, providing insights that guide research, policy, and practice (1). However, conducting systematic reviews or meta-analyses is an inherently complex and time-intensive process (2,3). Researchers and reviewers must meticulously find, read, and evaluate numerous articles—often titles, abstracts, and full texts—in multiple iterative rounds. Each round serves distinct purposes, such as relevance screening, quality assessment, or data extraction, frequently involving multiple reviewers to ensure reliability and mitigate bias. The need for such rigor, combined with the ever-growing volume of scientific publications, has made literature reviews dauntingly labor-intensive.

Compounding this challenge is the dynamic nature of systematic review protocols. Slight changes in the study design, such as updated inclusion criteria or expanded search strategies, often necessitate redoing or augmenting prior reviews. Such revisions can exponentially increase the effort required, making these studies resource-intensive and, at times, prohibitive. This is especially significant in fields like medicine, where systematic reviews often underpin evidence-based guidelines and decision-making. Hence, there is an urgent need for tools that not only accelerate these processes but also enhance their efficiency and reliability.

## 1.2. Purpose of the Package

LatteReview is a Python-based framework designed to address these challenges by automating key elements of the academic and systematic review process using advanced large language models (LLMs). Its name symbolizes the vision of transforming the traditionally arduous task of academic reviewing into an efficient and even enjoyable experience—*as quick and satisfying as savoring a latte*. By integrating multi-agent systems powered by LLMs, LatteReview enables researchers to perform complex review tasks with minimal effort while maintaining high standards of rigor and transparency.

The package leverages LLMs' capabilities to analyze and synthesize textual and visual data, making it possible to automate tasks like relevance scoring, conceptual abstraction, and structured data extraction. These tasks, traditionally performed by human reviewers, are accelerated and systematized using AI agents. Moreover, LatteReview's modular design supports workflows involving sequential or parallel review rounds, dynamic decision-making, and the ability to incorporate reviewer feedback for iterative refinement.

## 1.3. Audience and Use Cases

LatteReview is intended for a broad audience spanning academia, industry, and beyond. Its primary users are researchers, systematic reviewers, and data scientists who seek to streamline their literature review processes. Key use cases include:

- Systematic Reviews and Meta-Analyses: Automating title and abstract screening, checking against inclusion and exclusion criteria, relevance scoring, and data abstraction for large datasets.

- Rapid Reviews: Enabling quick assessments of recent literature, such as understanding the new trends in research or collecting new clinical guidelines or policy briefs.
- Preliminary Screening: Filtering articles based on defined criteria before conducting in-depth analyses.
- Interdisciplinary Collaborations: Facilitating reviews in emerging or interdisciplinary fields where traditional expertise may be limited.

LatteReview supports a wide array of inputs, including article titles, abstracts, and multimodal data (e.g., images). Its compatibility with various LLM providers ensures users can tailor the framework to their specific needs, from leveraging state-of-the-art cloud-based models to employing locally hosted solutions. With features such as Retrieval-Augmented Generation (RAG) for context integration (4), LatteReview enhances both the efficiency and depth of systematic review, setting a new standard for how research synthesis is conducted.

## 2. System Architecture and Design

### 2.1. High-Level Overview

The LatteReview framework is built around a multi-agent architecture designed to streamline systematic review workflows. At its core, the system integrates large language models (LLMs) with a structured workflow engine to provide an end-to-end solution for literature review tasks. The architecture is modular, enabling extensibility and customization to meet the unique demands of various research contexts.

The system comprises three primary components:

1. **Providers**: These interface with different LLM Application Programming Interface (APIs), including *OpenAI*, *Ollama*, and *LiteLLM*. Providers abstract the complexities of API interactions, ensuring uniformity and reliability across diverse model architectures.
2. **Reviewer Agents**: Representing the core review units, reviewer agents encapsulate different roles and expertise levels. They perform tasks such as relevance scoring, abstraction, and customized reviews based on user-defined schemas.
3. **Workflows**: This module orchestrates multi-round reviews, allowing agents to interact and process data sequentially or in parallel. Workflows manage dependencies, facilitate dynamic decision-making, and ensure structured output.

A high-level flow diagram illustrating the relationships between these components is provided below in Figure 1.

### 2.2. Core Modules

#### 2.2.1. Providers

The Provider module acts as the bridge between the LatteReview framework and LLM APIs. It includes:

- **BaseProvider**: An abstract class defining the core interface for all providers. It ensures consistent error handling, type safety, and API interaction.
- **OpenAIProvider**: Facilitates integration with OpenAI and Gemini models, supporting text and multimodal inputs.
- **OllamaProvider**: Optimized for local LLMs, enabling efficient processing without reliance on cloud services.
- **LiteLLMProvider**: A unified interface supporting multiple model providers, including OpenAI and Anthropic.

#### 2.2.2. Agents

Agents are specialized units within the framework responsible for executing review tasks. The following agent types are implemented:

1. **BaseReviewer**: A generic class that provides foundational capabilities for all agents, such as prompt handling, input validation, and output formatting.
2. **ScoringReviewer**: Designed for scoring-based evaluations, this agent generates structured outputs with scores, reasoning, and certainty metrics.
3. **TitleAbstractReviewer**: Mimics the ScoringReviewer agent but is specialized for reviewing input items (typically, titles and abstracts of some articles) against a provided inclusion and exclusion criteria.
4. **AbstractionReviewer**: Extracts structured data from unstructured inputs, leveraging predefined keys and detailed task-specific instructions.
5. **CustomReviewer**: A flexible template allowing users to define agents tailored to their unique requirements.

Each agent is instantiated with a specific provider, ensuring compatibility with the underlying LLM APIs. Agents use Pydantic-based validation to guarantee structured inputs and outputs, improving reliability and interpretability.

#### 2.2.3. Workflows

The Workflow module manages the orchestration of review tasks across agents and rounds. A review workflow accepts an input Pandas dataframe including the raw data to be reviewed, and returns an enhanced version of that dataframe with all review results added as new columns. Key features include:

- **Concept of Rounds**: Each round represents a stage in the review process, such as initial screening or expert validation.
- **Chaining Reviews**: Outputs from all agents in one round can serve as inputs to agents in subsequent rounds, enabling complex, multi-step analyses.
- **Parallel Reviews**: Allows multiple agents to operate concurrently in a single round, significantly improving throughput for large datasets.

- **Dynamic Filters**: Enables conditional processing, such as reviewing only items with conflicting scores from previous rounds.

### 2.3. Key Design Decisions

LatteReview is designed to seamlessly integrate its components, providing a scalable and efficient platform for academic literature reviews. The following key design principles guide the framework's architecture:

1. **Code Simplicity**: By leveraging Pydantic for input validation and structured output, the framework ensures ease of use and extensibility while maintaining robust error handling.
2. **Structured Outputs**: Outputs are formatted as standardized JSON, enhancing interpretability and integration with downstream workflows.
3. **Extensibility**: Users can define custom agents and workflows, ensuring the framework adapts to evolving research needs.
4. **Compatibility**: Broad support for popular LLM APIs, including cloud-based and locally hosted models, ensures flexibility.
5. **Multimodal Support**: The architecture accommodates diverse input types, such as text and images, enabling comprehensive analyses.
6. **Retrieval-Augmented Generation (RAG)**: Incorporates support for dynamic context retrieval to enhance accuracy and relevance.
7. **Asynchronous Operations**: The use of asynchronous programming improves processing speed, enabling efficient handling of large-scale datasets.

These design principles collectively contribute to LatteReview's effectiveness in automating and accelerating systematic review processes while maintaining the rigor and adaptability required in diverse research contexts.

## 3. Core Functionalities

The LatteReview package provides a robust set of functionalities that empower users to conduct complex review workflows leveraging AI-driven multi-agent systems. These core functionalities are designed to maximize flexibility, accuracy, and scalability in processing various types of inputs, including text and images. Below, we outline the primary features and capabilities of LatteReview.

### 3.1. Scoring Reviews

Scoring reviews are one of the foundational capabilities of LatteReview. With the *ScoringReviewer* class, users can configure AI agents to assess and assign scores to input items based on predefined criteria. Each scoring review is governed by:

- **Scoring Task**: A descriptive goal that the reviewer must accomplish (e.g., "Evaluate relevance to AI in radiology").
- **Scoring Set**: A defined range of allowable scores, ensuring consistency in evaluations (e.g., [1, 2, 3, 4, 5]).
- **Scoring Rules**: Specific guidelines to ensure consistency and reliability in evaluations (e.g., a scale of 1 to 5).
- **Reasoning Transparency**: The agent provides reasoning for its decisions, which can be configured to be brief or detailed depending on the ReasoningType selected. The reasoning cannot be None for the ScoringReviewer. It defaults to "brief" if not provided.
- **Certainty Scores**: The agent outputs a certainty score, quantifying the confidence in its assessment. These scores can later be used for calibration of agents, e.g., through conformal prediction setups, to improve trust in agent outputs.
- **Additional Context**: Users can provide additional context to enhance the input for the agent. This can be a string or a function dynamically called on every input to retrieve or generate context, such as integrating data from a retrieval-augmented generation (RAG) pipeline.

This functionality is ideal for tasks such as quality assessments, relevance scoring, or prioritization of items.

```python
from lattereview.providers import LiteLLMProvider
from lattereview.agents import TitleAbstractReviewer
from lattereview.workflows import ReviewWorkflow
import pandas as pd
import asyncio
from dotenv import load_dotenv

# Load environment variables from the .env file in the root directory of your project
load_dotenv()

# First Reviewer: Conservative approach
reviewer1 = TitleAbstractReviewer(
    provider=LiteLLMProvider(model="gpt-4o-mini"),
    name="Alice",
    backstory="a radiologist with expertise in systematic reviews",
    inclusion_criteria="The study must focus on applications of artificial intelligence in radiology.",
    exclusion_criteria="Exclude studies that are not peer-reviewed or not written in English.",
    model_args={"temperature": 0.2},
)

# Second Reviewer: More exploratory approach
reviewer2 = TitleAbstractReviewer(
    provider=LiteLLMProvider(model="gemini/gemini-1.5-flash"),
    name="Bob",
    backstory="a computer scientist specializing in medical AI",
    inclusion_criteria="The study must focus on applications of artificial intelligence in radiology.",
    exclusion_criteria="Exclude studies that are not peer-reviewed or not written in English.",
    model_args={"temperature": 0.2},
)

# Expert Reviewer: Resolves disagreements
expert = TitleAbstractReviewer(
    provider=LiteLLMProvider(model="gpt-4o"),
    name="Carol",
    backstory="a professor of AI in medical imaging",
    inclusion_criteria="The study must focus on applications of artificial intelligence in radiology.",
    exclusion_criteria="Exclude studies that are not peer-reviewed or not written in English.",
    model_args={"temperature": 0.2},
    additional_context="""
    Alice and Bob disagree with each other on whether or not to include this article.
    You can find their reasonings above.
    """,
)

# Define workflow
workflow = ReviewWorkflow(
    workflow_schema=[
        {
            "round": 'A',  # First round: Initial review by both reviewers
            "reviewers": [reviewer1, reviewer2],
            "text_inputs": ["title", "abstract"]
        },
        {
            "round": 'B',  # Second round: Expert reviews only disagreements
            "reviewers": [expert],
            "text_inputs": ["title", "abstract", "round-A_Alice_output", "round-A_Bob_output"],
            "filter": lambda row: row["round-A_Alice_evaluation"] != row["round-A_Bob_evaluation"]
        }
    ]
)

# Load and process your data
data = pd.read_excel("articles.xlsx")  # Must have 'title' and 'abstract' columns
results = asyncio.run(workflow(data))  # Returns a pandas DataFrame with all original and output columns

# Save results
results.to_csv("review_results.csv", index=False)
```

**Figure 2.** An example workflow for systematic review of some articles using the TitleAbstractReviewer agents.

## 3.2. Title and Abstract Reviews

The Title and Abstract Review functionality is a specialized extension of the ScoringReviewer class. While leveraging the core scoring mechanisms, it focuses specifically on evaluating titles and abstracts of research items based on predefined inclusion and exclusion criteria. This functionality provides a structured and efficient approach for screening research documents, ensuring that only the most relevant items are included in downstream workflows (**Figure 2**). The key elements of Title and Abstract Reviews are outlined below:

**Inclusion and Exclusion Criteria:**

- Users define clear inclusion and exclusion criteria to guide the review process.
- The agent assesses whether the provided title and abstract meet all inclusion criteria and violate none of the exclusion criteria.

**Categorized Scoring Rules:**

- Scoring rules are categorized into inclusion and exclusion criteria, ensuring consistency and clarity in evaluations.
- Scores are assigned on a 5-point scale:
  - **1:** Absolutely exclude.
  - **2:** Better to exclude.
  - **3:** Not sure (ambiguous).
  - **4:** Better to include.
  - **5:** Absolutely include.

**Reasoning Transparency:**

- The agent provides reasoning for its decisions, which could be "brief" or "cot", where the former means the agent should return a brief reasoning in 1 - 2 sentences, and the latter means a more detailed step-by-step reasoning is expected in the Chain-of-thought format.
- Similar to ScoringAgent, the reasoning cannot be None for the TitleAbstractReviewer. It defaults to "brief" if not provided.

**Customizable Prompts:**

- Users can provide additional context or examples to tailor the evaluation process to their specific needs.

This functionality is particularly well-suited for systematic literature reviews, meta-analyses, and other research workflows that require efficient and consistent screening of large datasets. By ensuring rigorous adherence to inclusion and exclusion criteria, the Title and Abstract Review feature enhances the reliability and accuracy of research workflows, paving the way for more focused and impactful analyses.

## 3.3. Abstraction Reviews

The *AbstractionReviewer* class is another reviewer class tailored for structured information extraction. By defining abstraction keys and corresponding descriptions, users can instruct the AI to extract specific data points from complex inputs. Key features include:

- **Key-Value Pair Extraction**: Extract data in a structured format, ensuring consistency and reliability.
- **Custom Instructions**: Provide detailed guidance for each abstraction key, enabling nuanced extraction tailored to user needs.
- **Contextual Support**: Include additional contextual information to refine the extraction process.
- **Additional Context**: Similar to scoring reviews, users can provide additional context as a string or function, dynamically enhancing inputs with external data.

This functionality is particularly useful for tasks like summarizing research papers, analyzing trends in the literature, extracting metadata, or generating structured abstracts from text (**Figure 3**).

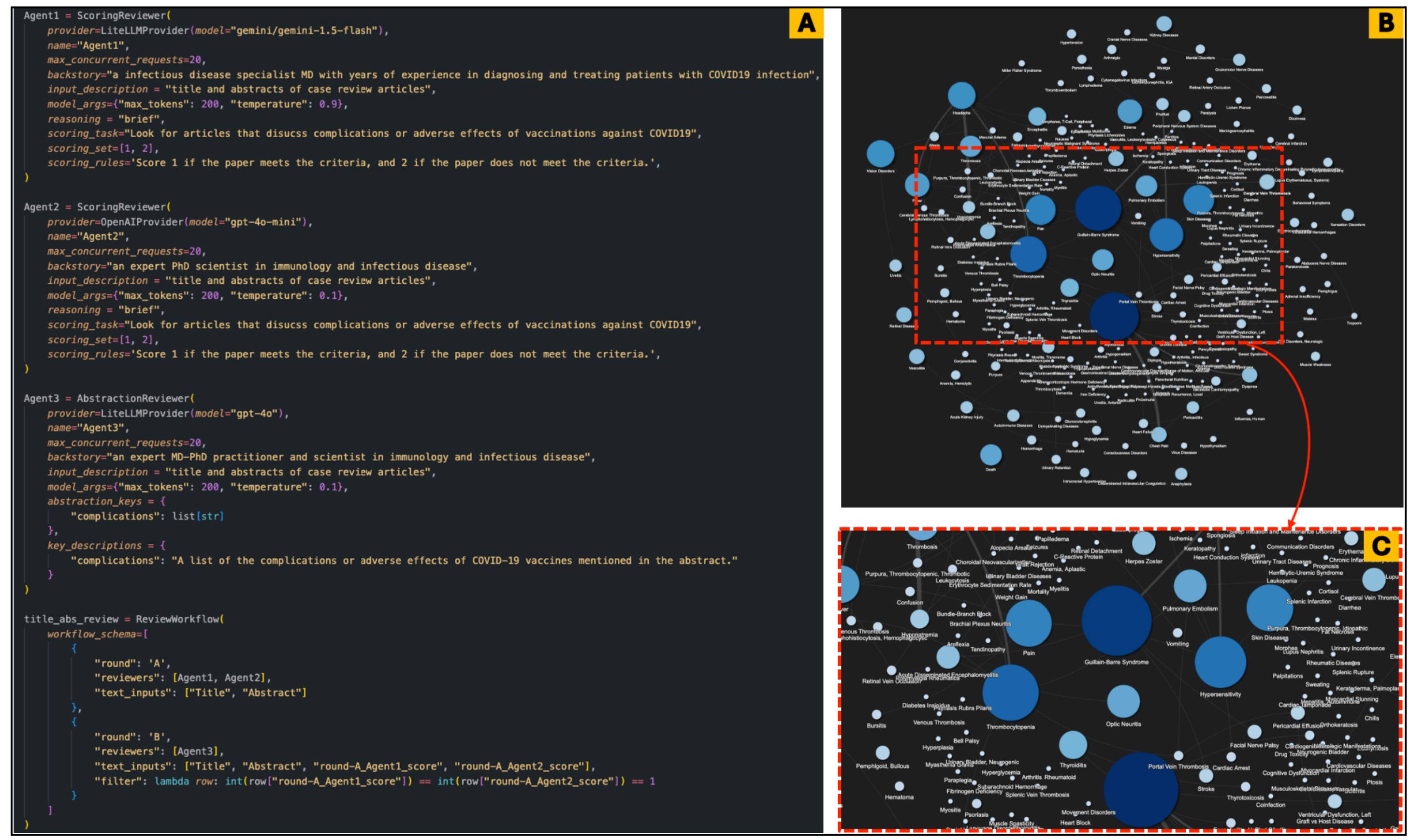

**Figure 3.** (A) A review workflow for screening titles and abstracts from a dataset created by querying PubMed for case report articles on complications of COVID-19 vaccines. The workflow consists of two agents that detect relevant articles and a third agent that abstracts the complications mentioned in those articles into a list. (B and C) A knowledge graph of complications extracted by the reviewers, where each node represents a complication. The size of the nodes correlates with the number of articles mentioning that complication, with larger and darker nodes indicating higher mention counts.

### 3.4. Multi-Reviewer Workflows

LatteReview excels in orchestrating workflows involving multiple AI reviewers. The *ReviewWorkflow* class allows users to define multi-stage review processes with:

- **Sequential Rounds**: Execute reviews in predefined stages, where the output of one stage can feed into the next.
- **Parallel Reviews**: Assign multiple reviewers to independently assess the same inputs in a given stage.

- **Conditional Filtering**: Define filters to determine which items proceed to subsequent rounds, enabling targeted and efficient workflows.
- **Output Aggregation**: Combine results from multiple reviewers, supporting advanced use cases like consensus-building or conflict resolution.

The multi-reviewer workflow capability is ideal for hierarchical decision-making processes and collaborative reviews.

### 3.5. Multimodal Review Support

LatteReview supports multimodal inputs, allowing reviewers to analyze both text and image data. This capability enables:

- **Text and Image Integration**: Reviewers can consider text descriptions alongside associated images, such as figures or charts.
- **Validation**: Ensure all image inputs meet format and existence requirements before processing.
- **Unified Outputs**: Combine insights from text and image reviews into a cohesive result.

This feature is particularly beneficial for domains like medical imaging or research where visual data complements textual information.

### 3.6. Retrieval-Augmented Generation (RAG) Integration

LatteReview integrates seamlessly with retrieval-augmented generation workflows, enabling reviewers to:

- **Leverage External Knowledge**: Fetch relevant information dynamically to enhance review accuracy.
- **Dynamic Context Addition**: Automatically incorporate retrieved data for each input item into the review process, improving relevance and grounding.
- **Customizable Retrieval Functions**: Support for user-defined functions to tailor retrieval logic to specific use cases.

RAG integration is especially powerful for tasks requiring dynamic knowledge updates, such as literature reviews or policy compliance checks.

### 3.7. Custom Agents

Last but not least, LatteReview's extensible architecture allows users to define custom reviewer agents by subclassing the BasicReviewer. This enables:

- **Tailored Prompts**: Create specialized prompts that address unique review requirements.
- **Custom Logic**: Override default behaviors to implement domain-specific functionality.
- **Dynamic Adaptability**: Integrate novel data formats or workflows seamlessly.
- **Additional Context**: Like other agents, custom agents can utilize additional context to enhance their performance, using either static strings or dynamic functions.

Examples of custom agents include sentiment analysis reviewers, compliance checkers, or creative text generators.

## 4. Evaluation

Evaluating LatteReview presents unique challenges due to the framework's inherent flexibility and the virtually infinite possibilities for constructing review workflows. Given a specific review task, users can implement vastly different approaches—from deploying a single reviewer agent to orchestrating complex workflows involving multiple agents working in parallel or series. For instance, when screening titles and abstracts from a dataset, one researcher might opt for a straightforward single-agent approach, while another might design an intricate workflow with five or six reviewer agents collaborating in various configurations. These different approaches inevitably yield varying performance metrics and computational costs. Furthermore, since LatteReview allows users to choose from a wide range of LLMs as the core engines for reviewer agents, evaluating the framework's performance becomes intertwined with assessing the capabilities of these underlying models in scientific review tasks. The evaluation of LLMs themselves remains a complex challenge in the field, with ongoing debates about appropriate benchmarks and metrics—a comprehensive discussion of which lies beyond the scope of this technical report.

| Article Abbreviation | Inclusion Criteria | Exclusion Criteria |
| --- | --- | --- |
| Appenzeller-Herzog_2019 (5) | - Patients with Wilson's Disease of any age or stage<br>- Study drug has to be one of four established therapies, namely DPen, trientine, TTM or Zn.<br>- Control could be placebo, no treatment or any other treatment that does not include the respective study drug<br>- Concomitant therapies had to be identical in the compared treatment arms<br>- Combination therapy regimens that include the respective monotherapy drug are not considered<br>- Prospective or retrospective studies reported<br>- Randomized, non-randomized controlled trials and comparative observational studies | - Animal studies, case reports, case series, cross-sectional studies, before-after studies, reviews, letters, abstract-only publications, editorials, diagnostic or other testing studies and non-controlled studies |
| Donners_2021 (6) | - Emicizumab studies providing:<br>1. Data on humans<br>2. Original PK data or modeled PK data or PK/PD relationships<br>3. Access to the abstract and the full text in English<br>- In the event of doubt regarding eligibility, the records or articles should be included. | Not specified |
| Jeyaraman_2020 (7) | - Studies are included if they meet the following PICOS criteria:<br>Population: Patients with knee osteoarthritis<br>Intervention: MSC therapy<br>Comparator: Usual care<br>Outcomes: Visual Analog Score (VAS) for Pain, Western Ontario McMaster Universities Osteoarthritis Index (WOMAC), Lysholm Knee Scale (Lysholm), Whole-Organ Magnetic Resonance Imaging Score (WORMS), Knee Osteoarthritis Outcome Score (KOOS), and adverse events<br>Study Design: Randomized controlled trials | - Observational studies and interventional studies without a comparator group<br>- Animal studies involving stem cell therapy for knee osteoarthritis models<br>- Reviews |
| Meijboom_2021 (8) | - Articles are included if they meet the following criteria:<br>1. Study involved transitioning from a TNF-alpha inhibitor (including etanercept, infliximab, and adalimumab) originator to a biosimilar<br>2. The number of patients who retransitioned is reported or can be calculated<br>3. The article is an original research article published in a peer-reviewed journal<br>4. The article included baseline characteristics of the patients who transitioned<br>5. The article is written in English<br>- Transitioning is defined as patients in whom the biosimilar was introduced after the originator, without treatment with other drugs in between. Retransitioning is defined as restarting the originator directly after discontinuing a biosimilar, without treatment with other drugs in between.<br>- Both transitioning and retransitioning involve changes with the same active biological substance. | Not specified |
| Muthu_2021 (9) | - To be included, a study should meet the following criteria:<br>1. The study should be an RCT with 1:1 parallel two-arm design<br>2. The study must be related to spine surgery involving preoperative or intraoperative or postoperative variables<br>3. The study must have a dichotomous primary or secondary outcome. | - Studies not involving human subjects<br>- Studies with continuous variable outcomes like pain scores, Oswestry Disability Index scores, time to union without predefined clinical success criteria<br>- Studies that did not report a statistically significant primary or secondary outcome measure |
| Oud_2018(10) | - Randomized controlled trials (RCT)s on four specialized psychotherapies (DBT: dialectic behavior therapy, MBT: mentalization-based treatment, TFP: transference-focused therapy and ST: schema therapy) for adults (18 years and older) with Borderline personality disorder (BPD), which includes an individual psychotherapy component and had a duration of 16 weeks or more.<br>- Eligible comparison groups are other protocolized and specialized psychotherapies, or control groups, for example, treatment as usual (TAU), waiting list, attention control or community treatment by experts (CTBE). | - Studies are excluded with a cut-off of <66% of the participants having BPD, unless disaggregated data are provided.<br>- Studies are excluded that tested incomplete versions of specialized treatment, for example, studies that investigated only skills training instead of the full DBT program. |

**Table 1.** Inclusion and exclusion criteria documented for six of the articles used to evaluate LatteReview from the Synergy collection of datasets.

Despite these methodological challenges, we aim to demonstrate LatteReview's practical utility in systematic review applications through two distinct evaluation approaches. First, we evaluate the framework using the SYNERGY collection, a comprehensive pool of article datasets compiled from previous systematic reviews, complete with their original inclusion and exclusion criteria. This dataset collection, generously assembled by De Bruin al.(11), provides a standardized benchmark for assessing systematic review automation tools. Second, we evaluate LatteReview on a custom dataset derived from two of our previously published scoping reviews on AI applications in cardiothoracic imaging (12,13). This second evaluation addresses a limitation we observed in the SYNERGY collection: the heterogeneity and occasional ambiguity in how inclusion and exclusion criteria were defined across its constituent datasets. While these criteria remain interpretable to human reviewers, they often lack the precision and structure that would optimize them for LLM comprehension. By including this custom dataset with carefully crafted prompts, we demonstrate LatteReview's performance under more idealized conditions.

For both evaluations, we implement a representative workflow utilizing the TitleAbstractReviewer agent from the LatteReview package. While not claiming to be optimal, this workflow serves as a practical starting point for academic reviews. The workflow comprises two junior reviewers—one powered by Google's Gemini-1.5-flash and another by OpenAI's GPT-4O-mini—both selected for their balance of performance and cost-effectiveness. A senior reviewer agent, utilizing OpenAI's GPT-4O, intervenes only in cases of disagreement between the junior reviewers to make final inclusion decisions. We note that alternative high-performance models such as OpenAI's O1, Claude 3.5 Sonnet, or Gemini 1.5 Pro could also serve effectively in the senior reviewer role.

The TitleAbstractReviewer agents in our workflow employ a 5-point Likert scale for article assessment, where a score of 1 indicates "absolutely exclude" and 5 indicates "absolutely include." For aggregating reviewer decisions, we implemented a hierarchical decision-making process: when the junior reviewers' scores differ or both assign a score of 3, the senior reviewer's assessment is sought and becomes definitive. In all other cases, we average the junior reviewers' scores to determine the final assessment.

| Dataset | Number of Articles (% Reported Relevant) | Sensitive Mode (T = 1.5) | Specific Mode (T = 3.0) | Balanced Mode (T = 4.5) | AUC |
|---|---|---|---|---|---|
| Oud 2018 | 952 (2.1) | Accuracy=0.63, Recall=1.0, Precision=0.05 | Accuracy=0.96, Recall=0.5, Precision=0.28 | Accuracy=0.93, Recall=0.8, Precision=0.21 | 0.95 |
| Appenzeller-Herzog 2019 | 2873 (0.9) | Accuracy=0.72, Recall=0.85, Precision=0.03 | Accuracy=0.98, Recall=0.23, Precision=0.12 | Accuracy=0.96, Recall=0.31, Precision=0.07 | 0.85 |
| Muthu 2021 | 2719 (12.36) | Accuracy=0.42, Recall=0.96, Precision=0.17 | Accuracy=0.84, Recall=0.18, Precision=0.27 | Accuracy=0.83, Recall=0.24, Precision=0.28 | 0.73 |
| Meijboom 2021 | 882 (4.2) | Accuracy=0.45, Recall=0.97, Precision=0.07 | Accuracy=0.89, Recall=0.84, Precision=0.26 | Accuracy=0.88, Recall=0.84, Precision=0.24 | 0.90 |
| Jeyaraman 2020 | 1175 (8.17) | Accuracy=0.85, Recall=0.5, Precision=0.27 | Accuracy=0.92, Recall=0.02, Precision=0.67 | Accuracy=0.92, Recall=0.02, Precision=0.67 | 0.71 |
| Donners 2021 | 258 (5.81) | Accuracy=0.13, Recall=1.0, Precision=0.06 | Accuracy=0.72, Recall=0.67, Precision=0.13 | Accuracy=0.68, Recall=0.73, Precision=0.12 | 0.77 |

**Table 2.** The performance of Review Workflow on six datasets from the Synergy Collection; AUC: Area Under the Receiver operating characteristic Curvel; T: Threshold

To comprehensively evaluate the framework's performance, we established three distinct decision thresholds, each representing different review strategies:

1. Sensitive Strategy (threshold = 1.5): Designed to maximize recall, this threshold includes articles with accumulated scores of 1.5 or higher, ensuring high sensitivity in capturing potentially relevant articles.
2. Specific Strategy (threshold = 4.5): Optimized for precision, this conservative threshold only includes articles that received very high eligibility scores, prioritizing specificity over sensitivity.
3. Balanced Strategy (threshold = 3.0): Represents a middle-ground approach, neither overly sensitive nor specific, aiming to balance precision and recall in article selection.

In the following sections, we present the framework's performance across both the SYNERGY dataset collection and custom dataset under each of these three review strategies, providing a comprehensive assessment of LatteReview's capabilities in systematic review automation.

## 4.1. Evaluation on SYNERGY Dataset Collection

**Table 1** presents a comprehensive summary of the datasets from the SYNERGY dataset collection utilized in our first evaluation, along with their respective inclusion and exclusion criteria. The performance metrics of our defined review workflow across these datasets are summarized in **Table 2.** A receiver operating characteristic (ROC) curve and a sensitivity-specificity plot for the performance are also visualized in **Figure 4**.

The review workflow demonstrated substantial discriminative capability, achieving Area Under the Curve (AUC) values ranging from 0.77 to 0.95 across the six different article collections from the SYNERGY dataset collection. However, despite these relatively high AUC values, we observed significant variation in performance across datasets. This variability can be attributed to several key factors:

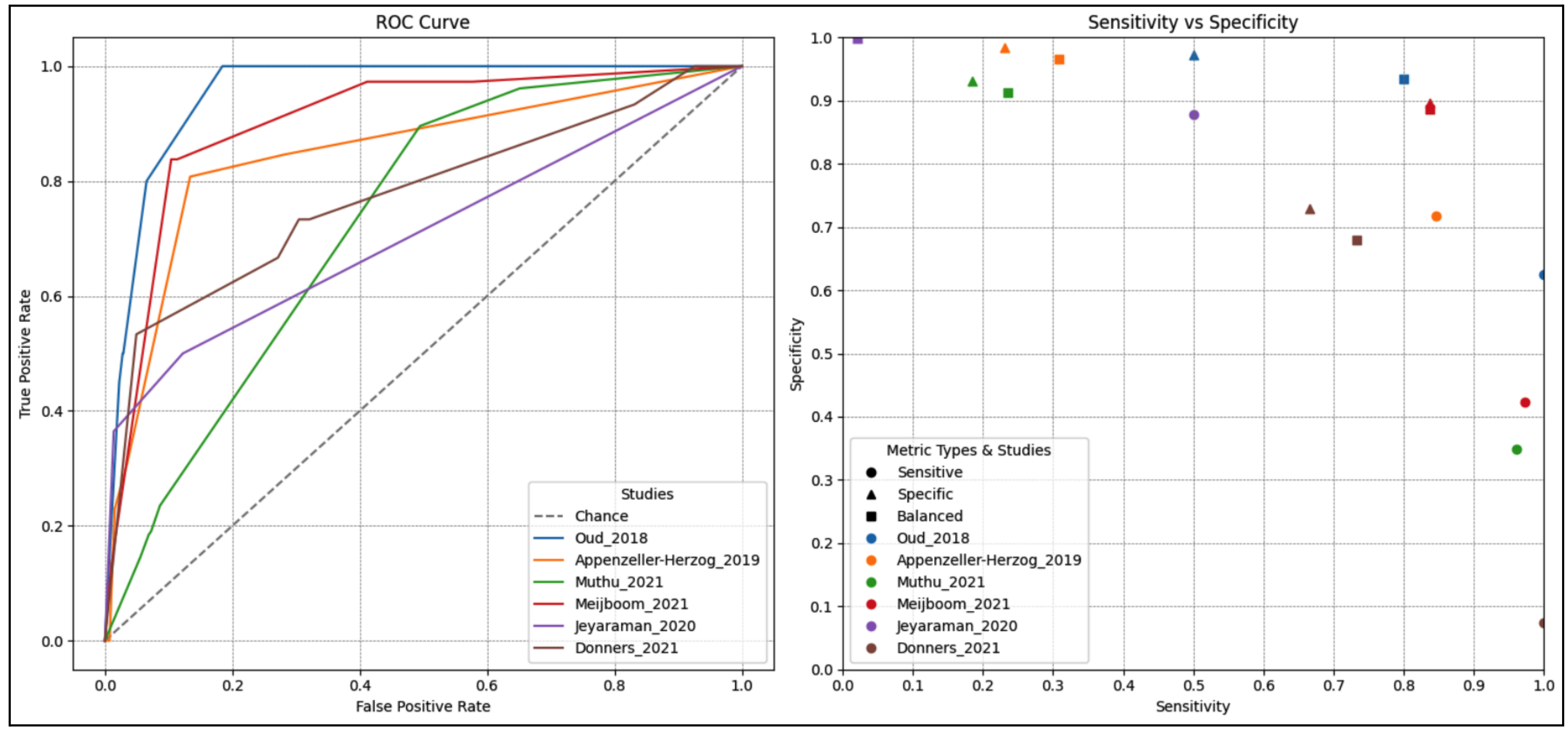

**Figure 4.** The performance of Review Workflow on six datasets from the Synergy Collection. The left plot shows the Receiver operating characteristic Curve for each dataset and the right plot demonstrates sensitivity vs. specificity of the Review Workflow in each performance mode.

1. Task Heterogeneity: The datasets encompass diverse review tasks, each with distinct inclusion and exclusion criteria that vary considerably in content, structure, clarity, and specificity. This heterogeneity in criteria definition directly impacts the LLMs' ability to interpret and apply the screening rules consistently.
2. Dataset Characteristics: The datasets exhibit substantial heterogeneity in both size and inclusion rates. Particularly noteworthy is the wide range in inclusion percentages, spanning from less than 1% to 12.36% of the total articles. In datasets with very low inclusion rates, the inclusion and exclusion criteria must be exceptionally specific and clearly articulated for LLMs to accurately identify the small subset of relevant articles.
3. Threshold Sensitivity: As illustrated in Figure 4, the choice of decision threshold significantly impacts the review workflow's performance. This threshold sensitivity highlights the importance of carefully calibrating the decision-making process based on the specific characteristics of each review task and the desired balance between precision and recall.We will further explore these performance dynamics in the subsequent section, where we discuss the value of validating the review workflow on a carefully selected subset of the datasets. This focused analysis provides additional insights into optimizing LatteReview's performance across different systematic review contexts.

| Search Strategy | Inclusion criteria | Exclusion criteria |
|---|---|---|
| Strategy 1 | 1. The study must involve CT scans. If multiple modalities are involved, CT scans should be among them. | 1. The study must not include PET scans as one of its modalities. |
| Strategy 2 | 1. The study must involve CT scans. If multiple modalities are involved, CT scans should be among them.<br>2. The study should introduce, develop, or discuss a deep learning-baseed classifier. | 1. The study must not include PET scans as one of its modalities.<br>2. Studies focusing on cardiovascular organs or lung vasculature must be excluded. |
| Strategy 3 | 1. The study must involve CT scans. If multiple modalities are involved, CT scans should be among them.<br>2. The study should introduce, develop, or discuss a deep learning-baseed classifier.<br>3. The clinical application or model in the study must focus on diagnosis. | 1. The study must not include PET scans as one of its modalities.<br>2. Studies focusing on cardiovascular organs or lung vasculature must be excluded.<br>3. All studies that did not have an external test set in addition to their internal test set must be excluded. |

**Table 3.** The three different search strategies put together for evaluating the Review Workflow on the custom dataset. The strategies get more complicated as we go from strategy 1 to 3.

### 4.2. Evaluation on Custom Dataset

In contrast to our first evaluation, which examined six distinct datasets from the SYNERGY collection, our second evaluation focuses on a single custom dataset derived from our previous publications. While utilizing a single dataset, we systematically assessed the LatteReview workflow's performance against three progressively complex pairs of inclusion and exclusion criteria. These criteria sets were designed to present increasing levels of complexity and comprehensiveness from pair 1 to pair 3.

**Table 3** presents these three distinct sets of inclusion and exclusion criteria, along with the number of eligible articles identified by human annotators under each criterion set. The performance metrics of the LatteReview workflow against these criteria sets are detailed in **Table 4** and visualized in **Figure 5**.

The analysis reveals several key insights:

1. Performance Range: The Area Under the Curve (AUC) for correctly identifying articles ranges from 0.79 for the most complex inclusion/exclusion criteria to 0.94 for the simplest criteria. While this range parallels that observed in our SYNERGY dataset collection evaluation, we note important qualitative differences in performance characteristics.
2. Decision Strategy Convergence: Notably, the balanced and specific review strategies yield similar performance metrics across all three reviewer configurations. This convergence suggests that the reviewer agents demonstrate high confidence in their scoring decisions, likely attributable to the clear and specific nature of the prompts derived from our carefully crafted inclusion and exclusion criteria.

3. Robust Performance Floor: Even in the most challenging configuration, the workflow maintains respectable performance metrics, with both sensitivity and specificity stabilizing around 0.7 when using the balanced threshold strategy. This performance stability is achieved without requiring extensive threshold validation or optimization, suggesting robust generalization capabilities.

It is worth noting that the workflow's performance could potentially be enhanced by employing more powerful LLMs for any of the three reviewer roles. The current results, achieved with relatively modest model configurations, demonstrate the framework's effectiveness when provided with well-structured, clear criteria—even when handling complex review tasks. On a separate note, the cost and time required for a review workflow can vary significantly depending on the selected base large language models and the tier of usage associated with each provider. However, in our experiments, a review workflow consisting of two junior reviewers using Gemini 1.5 Flash and OpenAI GPT-4O-mini with Tier 4 usage on OpenAI providers took, on average, about one minute to review 1,000 pairs of titles and abstracts, at a cost of $1.20.

For readers interested in reproducing these evaluations or examining the implementation details, we encourage accessing our package's GitHub repository[1], which contains the complete datasets, evaluation code, and workflow configurations used in these analyses.

| Search strategy | % Reported Relevant | Sensitive Mode (Threshold = 1.5) | Specific Mode (Threshold = 3.0) | Balanced Mode (Threshold = 4.5) | AUC |
|---|---|---|---|---|---|
| Strategy 1 | 5.62 | Accuracy=0.17, Recall=0.96, Precision=0.06 | Accuracy=0.7, Recall=0.78, Precision=0.13 | Accuracy=0.69, Recall=0.78, Precision=0.13 | 0.79 |
| Strategy 2 | 37.53 | Accuracy=0.45, Recall=0.99, Precision=0.4 | Accuracy=0.78, Recall=0.83, Precision=0.67 | Accuracy=0.78, Recall=0.84, Precision=0.66 | 0.82 |
| Strategy 3 | 71.27 | Accuracy=0.91, Recall=0.94, Precision=0.93 | Accuracy=0.91, Recall=0.9, Precision=0.97 | Accuracy=0.92, Recall=0.92, Precision=0.97 | 0.94 |

**Table 4.** The performance of Review Workflow on the custom dataset with three different search strategies (i.e., three different sets of inclusion and exclusion criteria that get more complicated from strategy 1 to 3); AUC: Area Under the Receiver operating characteristic Curve

## 5. Practical Tips

Reflecting on the evaluation results of LatteReview, we would like to provide users with some practical tips on how to effectively utilize this package and its functionalities. The following subsections outline key considerations and strategies for optimizing your use of LatteReview.

[1] https://github.com/PouriaRouzrokh/LatteReview

## 5.1. Design Review Workflows Thoughtfully

Building an ideal review workflow with LatteReview is akin to constructing a Lego masterpiece. LatteReview provides a set of core functionalities that researchers can combine to create optimal review workflows tailored to their specific use cases. Just as you would carefully design a review workflow with human reviewers, considering factors such as the number and roles of reviewers and the sequence of review rounds, apply the same principles when configuring LatteReview agents.

Ask yourself key questions: How many reviewers are needed for this task? What mix of junior and senior reviewers is optimal? At which stages should senior reviewers intervene? How will results from parallel junior reviewers be aggregated? By thoughtfully mapping out your review process as you would with human reviewers, you can effectively translate that design into a LatteReview workflow.

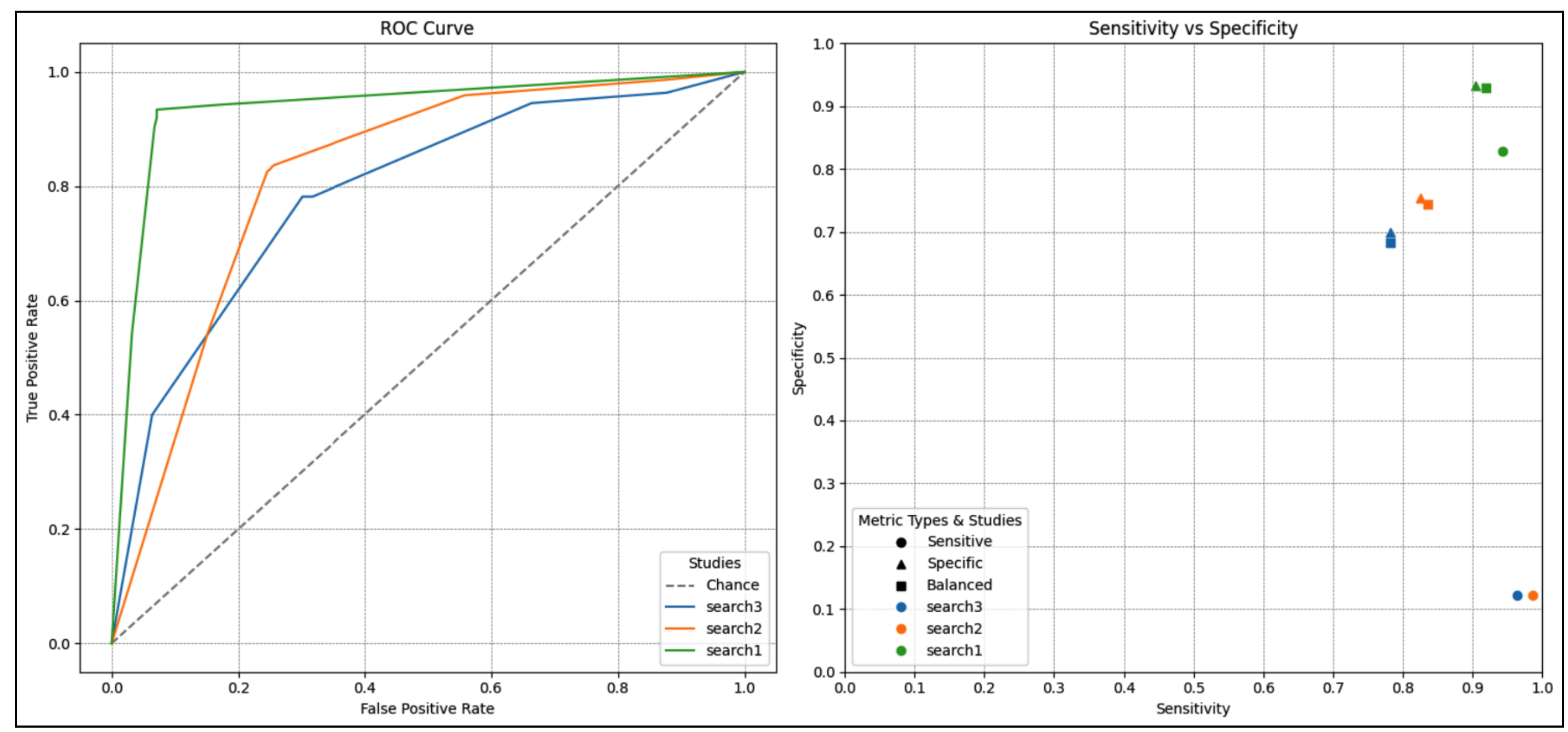

**Figure 5**. The performance of Review Workflow on the custom dataset with three different search strategies. The left plot shows the Receiver operating characteristic Curve for each search and the right plot demonstrates sensitivity vs. specificity of the Review Workflow in each performance mode.

## 5.2. Provide Clear Review Criteria and Instructions

Always define clear rules, instructions, or criteria for your review before configuring LatteReview agents. Providing vague or insufficient guidance will lead to suboptimal results, just as it would with human reviewers. This is especially critical for LLM-based reviewers, as these models rely on the prompts you provide to predict the most probable and relevant outputs.

The quality of your inclusion criteria, exclusion criteria, and other instructions directly impacts the performance of LatteReview agents. Invest time in carefully crafting comprehensive and unambiguous guidance to set your reviewers up for success. Well-defined prompts are the foundation upon which LLMs can generate accurate and useful review outputs.

## 5.3. Select Appropriate Reviewer Models

LatteReview allows you to initialize reviewer agents with different underlying LLM models, each with its own strengths, weaknesses, and costs. Always strive to match the right reviewer model to the right review task. Using an overly powerful and expensive model for a straightforward review is inefficient, while deploying an undersized model for a complex, nuanced review will yield poor results.

LatteReview offers access to a wide range of models from various LLM providers. Take advantage of this flexibility to create cost-effective workflows that pair the right models with the right tasks. For example, you might use smaller, more affordable models from the Gemini family for junior reviewers, a locally-hosted model for another junior reviewer, and a high-end OpenAI model with inference-time reasoning for your expert reviewer.

The optimal combination and hierarchy of reviewers depend on your specific use case. As a general heuristic, reserve more powerful and costly models for senior reviewer agents that are invoked when junior reviewers encounter difficulties or disagree. This allows you to deploy your LLM resources strategically and economically.

## 5.4. Enhance Reviewers with Examples and Context

View each reviewer agent as an LLM adapted for a specific review use case. Just as you would employ techniques to optimize the performance of LLMs in other applications, you can use similar approaches to enhance your LatteReview agents.

Two key strategies are providing examples and additional context to reviewer agents. By including a few illustrative examples of challenging review items and your expected outputs, you create a few-shot learning environment that primes the model to handle the full breadth and nuance of the review task. This is analogous to training human reviewers with sample cases.

Similarly, equipping reviewers with additional context can significantly improve their performance. This context might be task-specific, such as reference documents, or dynamically retrieved for each review item using techniques like Retrieval-Augmented Generation (RAG). LatteReview allows you to provide this context as a text input or a function, offering flexibility in how you augment your reviewers' knowledge and capabilities.

## 5.5. Align Reviewer Types with Review Tasks

LatteReview provides three default reviewer agent types, each suited for different review scenarios. The TitleAbstractReviewer is designed for screening articles based on their titles and abstracts against inclusion and exclusion criteria. The ScoringReviewer is appropriate when you need more nuanced scoring than a 5-point Likert scale include/exclude decision, offering configurable score ranges and review rules. The AbstractionReviewer is built for extracting structured information from review items.

If your review task does not fit neatly into these categories, LatteReview allows you to define custom reviewer classes that inherit from the BasicReviewer agent. This flexibility empowers you to create bespoke agents with input arguments and outputs tailored to your specific use case.
You can also design creative review workflows that combine multiple reviewer types. For example, you might use TitleAbstractReviewers or ScoringReviewers for initial screening, followed by AbstractionReviewers to extract structured data from the selected items. Aligning the right reviewer types with each stage of your review process is key to an efficient and effective workflow.

### 5.6. Optimize Database Search Queries

Before implementing LatteReview workflows, it is crucial to develop well-crafted search queries when retrieving articles from bibliographic databases. Many screening tasks that might initially seem to require sophisticated AI-based review can actually be handled efficiently through careful query construction and database filtering. For instance, in our evaluation of the SYNERGY dataset collection, we observed that certain inclusion criteria, such as restricting to English-language publications, could be entirely addressed through database filters rather than requiring LLM-based screening.

Even more sophisticated filtering requirements, such as distinguishing between human and animal studies or identifying specific study designs, can often be accomplished using the advanced search features and controlled vocabularies (e.g., MeSH terms in MEDLINE) available in scientific databases like PubMed, Scopus, or Web of Science. By leveraging these built-in filtering capabilities, researchers can significantly reduce the complexity of the subsequent review task and allow LatteReview agents to focus on more nuanced aspects of the screening process that genuinely require AI-based decision making.

### 5.7. Validate Workflows on Small Sample Sets

When applying LatteReview to large-scale review tasks, we strongly recommend first validating the workflow on a smaller, randomly selected subset of the data. This preliminary validation serves multiple crucial purposes:

1. Criteria Assessment: By manually reviewing a sample of articles against your inclusion/exclusion criteria, you can assess the clarity and applicability of these criteria from a reviewer's perspective. If you encounter difficulties interpreting or applying certain criteria, it's likely that LLM-based reviewers will face similar or greater challenges. This insight can be invaluable for refining criteria before scaling up to the full dataset.
2. Threshold Calibration: As demonstrated in our evaluation results, when inclusion/exclusion criteria are well-defined and clear, reviewer agents tend to provide more consistent and confident scores. In such cases, selecting an appropriate decision threshold becomes more straightforward, often eliminating the need for extensive threshold optimization. However, if you find that a simple threshold (e.g., 3.0 for balanced strategy) does not yield satisfactory results on your validation sample, you can use this smaller dataset to calibrate the optimal threshold before proceeding with the full review.

3. Example Collection: Manual review of a sample provides real-world examples of borderline or challenging cases. These examples can be incorporated into reviewer agents' prompts, enhancing their understanding of the nuanced decisions required for your specific review task. Well-chosen examples are particularly valuable for helping LLMs understand edge cases and subtle distinctions in your inclusion/exclusion criteria.
4. Performance Estimation: The validation sample provides an early indication of the workflow's expected performance on your full dataset. This allows you to make informed decisions about whether to proceed with the current configuration or make adjustments to the workflow, such as employing more powerful LLMs or refining the criteria definitions.

This validation approach embodies the principle of "fail fast, learn quickly" in systematic review automation. By investing time in thorough validation on a small scale, researchers can significantly improve the efficiency and reliability of their full-scale review process.

### 5.8. Integrate Human Oversight

While LatteReview offers powerful automation capabilities, it's essential to recognize the limitations inherent in any AI-based system. Even the most advanced LLMs can struggle with certain review items due to factors like ambiguity, lack of context, or insufficient instructions.

Therefore, it's crucial to design your LatteReview workflows with human oversight in mind. Rather than aiming for a fully automated end-to-end process, consider breaking your review into stages punctuated by manual review. Human reviewers can assess the outputs of each automated stage, provide feedback to refine the model prompts, and handle edge cases that the AI struggles with.

One effective approach is to configure your initial reviewer agents to be highly sensitive, aiming to exclude as many irrelevant items as possible. Human reviewers can then focus their efforts on the more ambiguous or challenging items that remain. This hybrid workflow leverages the efficiency of AI while preserving the nuance and domain expertise of human judgment.

By thoughtfully integrating human and AI reviewers, you can create a review process that achieves the best of both worlds - the speed and scale of automation with the adaptability and insight of human intelligence. LatteReview provides the building blocks, but it's up to you as the researcher to architect a system that leverages those capabilities most effectively for your specific review needs.

## 6. Conclusions and Future Directions

LatteReview provides a meaningful step toward enhancing the academic literature review process through the application of AI. By enabling customizable multi-agent workflows and supporting various data modalities, the package aims to simplify and streamline review tasks while maintaining flexibility for diverse use cases. Its integration with multiple LLM providers and support for Retrieval Augmented Generation highlights a thoughtful approach to incorporating advanced AI capabilities. While the system demonstrates significant promise in reducing the time and effort required for complex reviews, it remains focused on offering tools that align with the practical needs of its users.

Future development will focus on expanding LatteReview's capabilities and accessibility. Priorities include supporting a broader range of language models, facilitating improved collaboration between agents and users, and enhancing memory and context management to refine decision-making processes. Simplifying usability through a no-code interface and providing more advanced validation tools are also key areas of interest. By pursuing these enhancements, LatteReview aspires to evolve into a widely applicable tool that balances innovation with usability, continuing to assist researchers in their academic endeavors.

# Supplements

## 8. Installation and Dependencies

### 8.1. Installation

LatteReview is designed to be straightforward to install, ensuring accessibility for both researchers and developers. The package is available on PyPI and can be installed using the following command:

```
pip install lattereview
```

For users requiring additional functionalities, LatteReview provides optional extras:

- Development tools:
  `pip install "lattereview[dev]"`
- Documentation tools:
  `pip install "lattereview[docs]"`
- All extras (includes development and documentation tools):
  `pip install "lattereview[all]"`

For users who wish to explore the latest features or contribute to development, LatteReview can be installed directly from the source code:

```
git clone https://github.com/PouriaRouzrokh/LatteReview.git
cd LatteReview
pip install .
```

For development setups with all optional dependencies:

```
pip install -e ".[all]"
```

To verify the installation:

```
import lattereview
print(lattereview.__version__)
```

### 8.2. Dependencies

LatteReview relies on a carefully selected set of Python libraries to ensure robust functionality and seamless integration with external APIs. The table below summarizes the dependencies, their versions, and their relevance to different installation types:

| Dependency | Version | Main Installation | Development Tools | Documentation Tools |
|---|---|---|---|---|
| litellm | >=1.55.2 | ✓ | | |
| openai | >=1.57.4 | ✓ | | |
| pandas | >=2.2.3 | ✓ | | |
| pydantic | >=2.10.3 | ✓ | | |
| python-dotenv | >=1.0.1 | ✓ | | |
| tokencost | >=0.1.17 | ✓ | | |
| tqdm | >=4.67.1 | ✓ | | |
| openpyxl | >=3.1.5 | ✓ | | |
| black | >=24.10.0 | | ✓ | |
| flake8 | >=7.1.1 | | ✓ | |
| ipykernel | >=6.29.5 | | ✓ | |
| matplotlib | >=3.9.4 | | ✓ | |
| networkx | >=3.2.1 | | ✓ | |
| pyvis | >=0.3.2 | | ✓ | |
| scikit-learn | >=1.6.0 | | ✓ | |
| mkdocs | >=1.5.0 | | | ✓ |
| mkdocs-material | >=9.0.0 | | | ✓ |
| mkdocstrings | >=0.24.0 | | | ✓ |
| mkdocstrings-python | >=1.7.0 | | | ✓ |

### 8.3. Environment Configuration

LatteReview requires API keys for accessing LLM providers. These keys can be configured using environment variables. The recommended approach is to use a .env file in the project directory:

```
OPENAI_API_KEY = "..."
ANTHROPIC_API_KEY = "..."
GEMINI_API_KEY = "..."
GROQ_API_KEY = "..."
...
```

Then, you should load the environment variables in your script:

```
from dotenv import load_dotenv
load_dotenv()
```

Alternatively, you can set environment variables directly using the export command in your terminal:

```
export OPENAI_API_KEY="your_openai_key"
export ANTHROPIC_API_KEY="your_openai_key"
```

This method sets the variables for the current session. To make them persistent across sessions, add the export commands to your shell configuration file (e.g., .bashrc or .zshrc) and reload the shell:

```
source ~/.bashrc
```

This configuration ensures seamless integration with external APIs, enabling out-of-the-box functionality for all supported models.

## 5. Usage Examples and Tutorials for LatteReview

Here, we will provide three examples of how LatteReview could be applied in systematic review with basic, intermediate, and advanced levels of complexity. For more examples, please visit our tutorial page.

### 5.1. Basic Example: Single Reviewer Workflow

This example demonstrates how to use LatteReview to conduct a basic review using a single AI reviewer. The review process evaluates how relevant a research article is to artificial intelligence in radiology.

```
from lattereview.providers import LiteLLMProvider
from lattereview.agents import ScoringReviewer
from lattereview.workflows import ReviewWorkflow
import pandas as pd
import asyncio
from dotenv import load_dotenv

# Load environment variables from .env file
load_dotenv()

# Create a single reviewer
reviewer = ScoringReviewer(
       provider=LiteLLMProvider(model="gpt-4o-mini"),
       name="Reviewer1",
       backstory="A radiologist with expertise in AI applications",
       scoring_task="Evaluate relevance of the article to AI in radiology",
       scoring_set=[1, 2, 3, 4, 5],
       scoring_rules="Rate relevance on a scale of 1 (not relevant) to 5 (highly relevant).",
       model_args={"temperature": 0.2}
)

# Define a workflow with the single reviewer
workflow = ReviewWorkflow(
       workflow_schema=[
       {
              "round": 'A',
              "reviewers": [reviewer],
```

```
            "text_inputs": ["title", "abstract"]
      }
      ]
)

# Load data
data = pd.DataFrame({
      'title': ['Advances in AI'],
      'abstract': ['A comprehensive review of recent developments in AI applications for
radiology.']
})

# Run the workflow
results = asyncio.run(workflow(data))

# Save results
results.to_csv("basic_review_results.csv", index=False)
Alternative Code Implementation
For simpler use cases, reviewers can directly process input strings without defining workflows:
# Directly use the reviewer for a list of inputs
inputs = [
      "Advances in AI in Radiology",
      "AI in Cardiology: A review of applications."
]

# Review items
results, costs = asyncio.run(reviewer.review_items(inputs))

# Display results
for input_text, result in zip(inputs, results):
      print(f"Input: {input_text}\nScore: {result['score']}\nReasoning:
{result['reasoning']}\n")
```

Explanation:

1. Provider Configuration: LiteLLMProvider is used to access the GPT-4 model.
2. Reviewer Creation: The ScoringReviewer is configured with specific rules for rating relevance.
3. Workflow Definition: A single-round workflow is created where the reviewer evaluates the relevance of titles and abstracts.
4. Direct Use: Reviewers can also directly process text lists for quick analyses.

### 5.2. Intermediate Example: Two Junior Reviewers and a Senior Reviewer

In this example, we introduce a second round where a senior reviewer resolves conflicts between two junior reviewers.

```
# Define junior reviewers
reviewer1 = ScoringReviewer(
      provider=LiteLLMProvider(model="gpt-4o-mini"),
```

```
        name="Alice",
        backstory="A radiologist focusing on systematic reviews",
        scoring_task="Evaluate relevance of the article to AI in radiology",
        scoring_set=[1, 2, 3, 4, 5],
        scoring_rules="Rate relevance from 1 (not relevant) to 5 (highly relevant).",
        model_args={"temperature": 0.1}
)

reviewer2 = ScoringReviewer(
        provider=LiteLLMProvider(model="gemini/gemini-1.5-flash"),
        name="Bob",
        backstory="A computer scientist specializing in AI for medicine",
        scoring_task="Evaluate relevance of the article to AI in radiology",
        scoring_set=[1, 2, 3, 4, 5],
        scoring_rules="Rate relevance from 1 (not relevant) to 5 (highly relevant).",
        model_args={"temperature": 0.8}
)

# Define senior reviewer
expert = ScoringReviewer(
        provider=LiteLLMProvider(model="gpt-4o"),
        name="Carol",
        backstory="A professor of AI in medical imaging",
        scoring_task="Resolve disagreements between Alice and Bob",
        scoring_set=[1, 2],
        scoring_rules="Score 1 if agreeing with Alice, 2 if agreeing with Bob",
        model_args={"temperature": 0.2}
)

# Define workflow
workflow = ReviewWorkflow(
        workflow_schema=[
        {
                "round": 'A',
                "reviewers": [reviewer1, reviewer2],
                "text_inputs": ["title", "abstract"]
        },
        {
                "round": 'B',
                "reviewers": [expert],
                "text_inputs": ["title", "abstract", "round-A_Alice_output", "round-A_Bob_output"],
                "filter": lambda row: row["round-A_Alice_score"] != row["round-A_Bob_score"]
        }
        ]
)

# Load and run data
data = pd.read_excel("articles.xlsx")
results = asyncio.run(workflow(data))
results.to_csv("intermediate_review_results.csv", index=False)
```

Explanation:

1. Conflict Resolution: Disagreements between junior reviewers are filtered for expert review in round B.
2. Parallel Reviews: Two reviewers operate concurrently in round A.
3. Filtering: The filter function identifies cases where the junior reviewers disagreed.
4. Output Columns: Results include outputs from all reviewers, enabling traceability of decisions.

### 5.3. Advanced Example: Retrieval Augmented Generation (RAG) with Custom Reviewer

This advanced example demonstrates the usage of LatteReview for integrating Retrieval Augmented Generation (RAG) into a multi-agent workflow, including a custom reviewer agent. The goal is to perform a comprehensive review of academic articles, utilizing additional context fetched dynamically from an external database and combining it with structured input processing.

```
from lattereview.providers import LiteLLMProvider
from lattereview.agents import ScoringReviewer, AbstractionReviewer, BasicReviewer
from lattereview.workflows import ReviewWorkflow
import pandas as pd
import asyncio
from dotenv import load_dotenv

# Load environment variables
load_dotenv()

# Define a function to provide additional context dynamically (RAG)
def retrieve_context(input_text):
      # Simulate a database or API call to fetch context
      return f"Additional insights about '{input_text}' fetched from external sources using
RAG."

# Custom Reviewer: Context-Aware Reviewer
class ContextAwareReviewer(BasicReviewer):
      generic_prompt = """
      **Analyze the input text with the following context:**
      <<${additional_context}$>>

      **Input Text:**
      <<${item}$>>

      **Task:**
      1. Provide a brief summary.
      2. Assess the relevance to artificial intelligence in radiology.

      **Output Format:**
      - `summary`: str
      - `relevance_score`: int (1 to 5)
      """
      response_format = {
```

```
        "summary": str,
        "relevance_score": int
        }
        input_description = "academic text with external context"

        def model_post_init(self, __context):
        try:
                self.setup()
        except Exception as e:
                raise Exception(f"Error initializing ContextAwareReviewer: {e}")

# Initialize Reviewers
scoring_reviewer_1 = ScoringReviewer(
        provider=LiteLLMProvider(model="gpt-4o-mini"),
        name="ScoringAgent1",
        scoring_task="Evaluate the relevance to AI in radiology",
        scoring_set=[1, 2, 3, 4, 5],
        scoring_rules="Rate relevance on a scale of 1 (low) to 5 (high).",
        model_args={"temperature": 0.1}
)

scoring_reviewer_2 = ScoringReviewer(
        provider=LiteLLMProvider(model="gemini/gemini-1.5-flash"),
        name="ScoringAgent2",
        scoring_task="Evaluate the relevance to AI in radiology",
        scoring_set=[1, 2, 3, 4, 5],
        scoring_rules="Rate relevance on a scale of 1 (low) to 5 (high).",
        model_args={"temperature": 0.8}
)

abstraction_reviewer = AbstractionReviewer(
        provider=LiteLLMProvider(model="gpt-4o"),
        abstraction_keys={"title": str, "abstract": str},
        key_descriptions={
        "title": "Extract the article title.",
        "abstract": "Extract the article abstract."
        }
)

context_aware_reviewer = ContextAwareReviewer(
        provider=LiteLLMProvider(model="gpt-4o"),
        name="ContextReviewer",
        additional_context=retrieve_context
)

# Define Workflow
workflow = ReviewWorkflow(
        workflow_schema=[
        {
                "round": 'A',
                "reviewers": [abstraction_reviewer],
```

```
            "text_inputs": ["raw_text"]
    },
    {
            "round": 'B',
            "reviewers": [scoring_reviewer_1, scoring_reviewer_2],
            "text_inputs": ["round-A_AbstractionReviewer_output"]
    },
    {
            "round": 'C',
            "reviewers": [context_aware_reviewer],
            "text_inputs": ["round-B_ScoringAgent1_output", "round-B_ScoringAgent2_output"],
            "filter": lambda row: row["round-B_ScoringAgent1_score"] > 3 and
row["round-B_ScoringAgent2_score"] > 3
    }
    ]
)

# Sample Data
data = pd.DataFrame({
    'raw_text': [
    "Exploring the role of AI in detecting lung cancer from radiographs.",
    "A systematic review of machine learning in cardiac imaging.",
    "The intersection of AI and pathology: Challenges and opportunities."
    ]
})

# Run the Workflow
results = asyncio.run(workflow(data))

# Save Results
results.to_csv("advanced_review_results.csv", index=False)
```

Explanation:

1. Custom Reviewer Integration:
    - The ContextAwareReviewer demonstrates extensibility by integrating external context dynamically fetched through a custom function.
    - Its generic_prompt is designed to combine the input text and the fetched context for comprehensive analysis.
2. Workflow Configuration:
    - Round A: The AbstractionReviewer extracts the title and abstract from the raw text.
    - Round B: Two ScoringReviewers independently evaluate the relevance of the extracted text to AI in radiology.
    - Round C: The ContextAwareReviewer uses both the outputs from Round B and additional context to provide a refined summary and relevance score.
    - Filtering: Round C processes only those articles which both Round B reviewers rated above 3.

3. RAG Integration:
    - The fetch_context function simulates a Retrieval Augmented Generation approach, dynamically augmenting the input with external insights.
4. Result Management:
    - Outputs from each round are systematically stored in columns, ensuring traceability and transparency.
    - The results DataFrame includes:
        - Extracted titles and abstracts (Round A).
        - Relevance scores and reasoning (Round B).
        - Context-aware summaries and relevance reassessment (Round C).